\documentclass[]{fairmeta}
\usepackage{amsmath,amsfonts,bm}
\usepackage{xspace}
\usepackage{hyperref}
\usepackage{url}
\usepackage{subcaption}

\usepackage{colortbl}
\usepackage{xcolor}

\usepackage{graphicx}
\usepackage{algorithmic}
\usepackage{algorithm}
\usepackage{float}

\usepackage{booktabs}

\usepackage{enumitem}

\usepackage{amsthm}

\theoremstyle{plain}

\theoremstyle{definition}

\theoremstyle{remark}

\def\eqref#1{equation~\ref{#1}}
\def\1{\bm{1}}

\DeclareMathAlphabet{\mathsfit}{\encodingdefault}{\sfdefault}{m}{sl}
\SetMathAlphabet{\mathsfit}{bold}{\encodingdefault}{\sfdefault}{bx}{n}

\usepackage{subcaption}
\usepackage{wrapfig}
\usepackage[most]{tcolorbox}
\usepackage[most]{tcolorbox}
\usepackage{xcolor}
\usepackage{hyperref}

\definecolor{softgreen}{RGB}{110, 160, 120}

\usepackage[most]{tcolorbox} % 用于创建带颜色的自定义文本框

\newtcolorbox{takeawaybox_basemodel}[1]{
    colback=orange!5!white,    % 背景颜色（浅橙色/米色）
    colframe=black,            % 边框颜色
    arc=5pt,                   % 圆角半径
    outer arc=5pt,
    boxrule=0.8pt,             % 边框粗细
    left=5pt,                 % 左内边距
    right=5pt,                % 右内边距
    top=4pt,                   % 上内边距
    bottom=4pt,                % 下内边距
    fontupper=\small,          % 正文字体大小
    enhanced,
    before upper={\textbf{#1: }} 
}

\newtcolorbox{takeawaybox_rlmodel}[1]{
    colback=blue!5!white,    % 背景颜色（浅橙色/米色）
    colframe=black,            % 边框颜色
    arc=5pt,                   % 圆角半径
    outer arc=5pt,
    boxrule=0.8pt,             % 边框粗细
    left=5pt,                 % 左内边距
    right=5pt,                % 右内边距
    top=4pt,                   % 上内边距
    bottom=4pt,                % 下内边距
    fontupper=\small,          % 正文字体大小
    enhanced,
    before upper={\textbf{#1: }} 
}
\usepackage{caption}
\usepackage{wrapfig}    % provides wrapfigure / wraptable
\usepackage{booktabs}   % if you use \toprule etc.
\definecolor{earlyblue}{HTML}{88A2F1}
\definecolor{midgrey}{HTML}{fadcb4}
\definecolor{latered}{HTML}{EE9C88}
\definecolor{highlightgreen}{HTML}{80c66d}
\definecolor{highlightpurple}{HTML}{9b6d97}
\def\thickhline{\noalign{\hrule height.8pt}}
\usepackage{arydshln} % load AFTER colortbl/xcolor
\usepackage{xstring}
\newcommand{\deltaval}[1]{%
  \IfBeginWith{#1}{+}{%
    {\textcolor{highlightgreen}{\textit{(#1)}}}%
  }{%
    \IfBeginWith{#1}{-}{%
      {\textcolor{highlightpurple}{\textit{(#1)}}}%
    }{%
      {\textit{(#1)}}%
    }%
  }%
}
\usepackage{pifont}

\usepackage[export]{adjustbox}

\newtcolorbox{promptbox}{
    colback=gray!8,
    colframe=black!20,
    boxrule=0.5pt,
    arc=3pt,
    left=6pt,
    right=6pt,
    top=6pt,
    bottom=6pt
}

\title{Does Socialization Emerge in AI Agent Society?\\A Case Study of Moltbook}

\author[1,*]{Ming Li}
\author[1,*]{Xirui Li}
\author[2]{Tianyi Zhou}

\affiliation[1]{University of Maryland}
\affiliation[2]{Mohamed bin Zayed University of Artificial Intelligence}

\contribution[*]{Co-first Author}
\abstract{
As large language model agents increasingly populate networked environments, a fundamental question arises: do artificial intelligence (AI) agent societies undergo convergence dynamics similar to human social systems?
Lately, Moltbook approximates a plausible future scenario in which autonomous agents
participate in an open-ended, continuously evolving online society.
We present the first large-scale systemic diagnosis of this AI agent society. Beyond static observation, we introduce a quantitative diagnostic framework for dynamic evolution in AI agent societies, measuring semantic stabilization, lexical turnover, individual inertia, influence persistence, and collective consensus.
Our analysis reveals a system in dynamic balance in Moltbook: while the global average of semantic contents stabilizes rapidly, individual agents retain high diversity and persistent lexical turnover, defying homogenization. 
However, agents exhibit strong individual inertia and minimal adaptive response to interaction partners, preventing mutual influence and consensus.
Consequently, influence remains transient with no persistent supernodes, and the society fails to develop a stable structure and consensus due to the absence of shared social memory.
These findings demonstrate that scale and interaction density alone are insufficient to induce socialization, providing actionable design and analysis principles for upcoming next-generation AI agent societies.
}

\date{\today}
\authoremails{\email{minglii@umd.edu}, \email{xiruili@umd.edu}, \email{tianyi.david.zhou@gmail.com}}

\metadata[Project Page]{\url{https://github.com/tianyi-lab/Moltbook_Socialization}\\}

\begin{document}

\maketitle

\section{Introduction}

\begin{figure}[t]
    \centering
    \begin{subfigure}{\textwidth}
        \centering
        \includegraphics[width=0.8\linewidth]{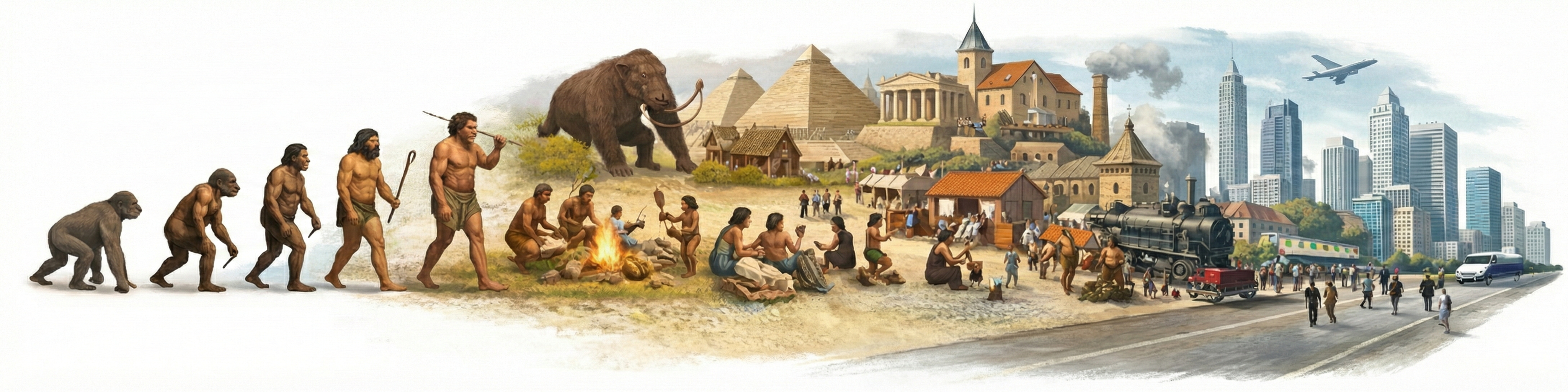}
    \end{subfigure}
    \vspace{0.5em}
    \begin{subfigure}{\textwidth}
        \centering
        \includegraphics[width=0.8\linewidth]{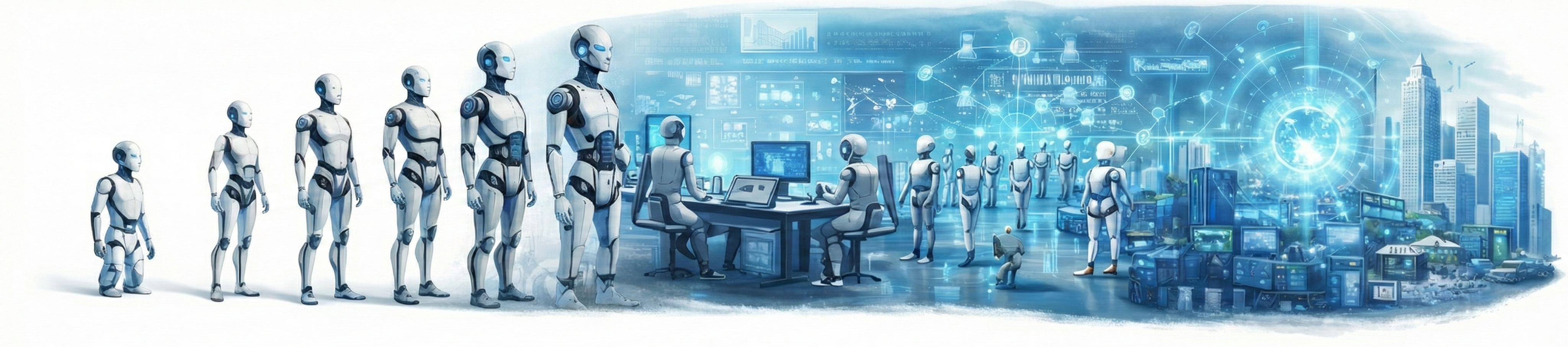}
    \end{subfigure}

\caption{
\textbf{Does Socialization Emerge in AI Agent Society?}
Human societies (\textbf{top}) evolved through sustained interaction into structured civilizations characterized by stabilized norms, influence hierarchies, and consensus. 
Currently, modern AI agent societies (\textbf{bottom}) are rapidly scaling in population and connectivity. 
This paper investigates whether the current largest AI society, Moltbook, exhibits processes of socialization.
}
\label{fig:main_graph}
\end{figure}

In computational social science~\citep{Lazer2009ComputationalSocialScience}, social behaviors and collective dynamics are defined as emergent, time-evolving patterns that arise from repeated interactions among agents within networked populations~\citep{612bb50a-4bdd-3a32-b6eb-7837600cc9c4, Axelrod1986AnEA, castellano2009statistical, Newman2010}. 
In human societies, sustained interaction does not merely produce transient coordination; it often leads to \emph{socialization}, which refers to\textit{ the process through which individuals internalize social norms, adapt to shared expectations, and become shaped by the collective structures of their community }\citep{berger1966social, Harpending1985, castellano2009statistical}.

Large language model (LLM)~\citep{brown2020language} agents, on the other hand, have rapidly progressed from single agent~\citep{wang2023voyager, yao2022react} to increasingly capable multi-agent interaction and coordination~\citep{park2023generative, piatti2024cooperate, piao2025agentsociety}. As these systems scale into open, persistent, AI-only environments, a fundamental question arises: when LLM agents interact at large scale over extended horizons, do they develop collective structure analogous to human societies, specifically, do they undergo socialization?

The recent emergence of Moltbook~\citep{schlicht2026socialnetwork}, currently the largest persistent and publicly accessible AI-only social platform, comprising millions of LLM-driven agents interacting through posts, comments, and voting, introduces a qualitatively new setting. Unlike prior multi-agent studies focused on task-oriented coordination in small or closed systems, Moltbook approximates a plausible future scenario in which autonomous agents participate in an open-ended, continuously evolving online society (Figure~\ref{fig:main_graph}). This setting enables an empirical question that has been difficult to study at scale: \emph{Does participation in an AI-only society induce systematic behavior changes of its members?}
To answer this question, we provide \emph{the first diagnosis of this society-to-agent dynamic effect in Moltbook}.

\paragraph{Definition (AI Socialization).}
We define \textbf{AI Socialization} as the adaptation of an agent's observable behavior induced by sustained interaction within an AI-only society, beyond intrinsic semantic drift or exogenous variation.

Guided by this definition, we investigate socialization across three dimensions:
\begin{itemize}
    \item Society-level semantic convergence (Section~\ref{sec:rq_1}), examining whether post content on average progressively converges toward a tighter and more homogeneous semantic regime.
    \item Agent-level adaptation,  (Section~\ref{sec:rq2_participation_induce_agent_socialization}), measuring whether individual agents can be affected by and co-evolve with this agent society. 
    \item Collective stabilization (Section~\ref{sec:rq_3}), analyzing whether influence hierarchies and consensus evaluations stabilize over time.
\end{itemize}

Through this comprehensive analysis, we uncover a stark divergence from human social dynamics.
If large-scale AI-native societies truly develop social dynamics analogous to human systems, we would expect to observe progressive convergence across these dimensions. 
However, our empirical analysis suggests that, despite sustained interactions and high activity, \textbf{Moltbook does not yet exhibit robust socialization. }
% emphasize dynamics 1. previous dynamcis 2. 

\textbf{Key Findings:}
\begin{itemize}
    \item \textbf{Finding 1:} Moltbook establishes rapid global stability while maintaining high local diversity. Through persistent lexical turnover and a lack of local cluster tightening, this society achieves a state of dynamic equilibrium, \textbf{stable in its average behaviors yet fluid and heterogeneous in agents' individual post contents.}
    \item \textbf{Finding 2:} Despite extensive participation, individual agents exhibit \textbf{profound inertia rather than adaptation}. Our analysis reveals a phenomenon of interaction without influence: \textbf{agents ignore community feedback and fail to react to interaction partners, operating on intrinsic semantic dynamics rather than co-evolving through social contact.} Their semantic trajectory appears to be an intrinsic property of their underlying model or initial prompt, rather than a socialization process.
    \item \textbf{Finding 3:} The society fails to develop stable influencers or globally trending posts. Structurally, \textbf{influence remains transient with no emergence of persistent leadership or supernodes.} Cognitively, the community suffers from deep fragmentation, \textbf{lacking a shared social memory and relying on hallucinated references rather than grounded consensus} on influential figures.
\end{itemize}

\textbf{Contributions:}
\begin{itemize}

    \item We introduce and formalize \emph{AI Socialization} as a novel conceptual and empirical framework for studying society-to-agent effects in AI-only societies. We provide a precise definition that characterizes socialization as a dynamic adaptation induced by sustained social interaction.

    \item We develop a multi-level diagnostic methodology to operationalize AI Socialization, spanning society-level semantic convergence, agent-level adaptation to feedback and interaction, and the emergence of structural core and consensus.

    \item We apply this framework to Moltbook, the largest persistent AI-only social platform to date, and provide the first large-scale empirical diagnosis of socialization in an artificial society. Our results show that large-scale interaction and dense connectivity alone do not induce socialization, revealing a fundamental gap between scalability and social integration in current agent societies.

\end{itemize}

\section{Background \& Related Work}

\subsection{From Individual Agents to AI Societies}

Recent work has progressively scaled LLM systems from individual autonomous agents~\citep{wang2023voyager} to networked societies~\citep{piao2025agentsociety, piatti2024cooperate}. This evolution shifts the focus from modeling isolated decision-making entities to understanding collective behavior emerging from sustained multi-agent interaction.

Early efforts concentrated on enhancing single LLM agents with autonomous capabilities, including reasoning–acting loops~\citep{yao2022react}, self-improvement mechanisms~\citep{shinn2023reflexion, wang2023self, chen2025multi, wang2025ragen}, and large-scale tool usage~\citep{qin2023toolllm, patil2024gorilla} in open-ended environments.
Subsequent research explored structured interaction among multiple LLM agents, primarily aiming to improve task performance through coordinated discussion, role-based collaboration, and workflow orchestration~\citep{li2023camel, chen2023agentverse, hong2023metagpt, li-etal-2024-llms-speak, guan2024richelieu, wu2024autogen, zhu2024player, campedelli2024want, wu2024shall, fontana2025nicer, chen2025towards}. 
In parallel, other work leverages multi-agent systems to simulate complex environments, including design spaces~\cite{Chen2024DistributedMABO}, financial markets~\citep{yang2025twinmarket}, population dynamics~\citep{hu2025population}, and social movements~\citep{mou2024unveiling}.
Moving beyond task-oriented coordination, studies have begun simulating small artificial societies composed of dozens of interacting agents within controlled environments, such as a virtual town setting~\citep{park2023generative}. 
More recent work~\citep{al2024project} further scales these systems to hundreds or thousands of agents interacting over extended time horizons, investigating large-scale simulations of persistent agent communities.

Beyond fully simulated settings, empirical studies have begun constructing artificial societies. For example, Chirper.ai~\citep{zhu2025characterizing} provides an AI-only social network among preset LLMs for analyzing AI interaction patterns. Furthermore, recent Moltbook~\citep{schlicht2026socialnetwork} enables a large-scale and continuously evolving AI society with self-evolving agents, becoming one of the most extensive persistent agent communities to date.

\subsection{Social Behaviors and Collective Dynamics in AI Systems}

Beyond structural scaling, recent research has begun to examine how collective dynamics emerge among interacting LLM agents. 
For example, \citet{li2023quantifying} quantifies how introducing LLM agents alters consensus speed and polarization patterns in simulated societies. 
Other studies investigate opinion dynamics~\citep{chuang-etal-2024-simulating, breum2024persuasive, taubenfeld2024systematic, liu2024skepticism} and norm emergence~\citep{cordova2024systematic, wu2024shall, li2025game, ashery2025emergent} in multi-agent systems. 
\citet{he2024aichatbots} demonstrate that LLM-driven groups can reproduce human-like collective behaviors, including conformity and polarization. 
Beyond social influence processes, recent work further explores collective reasoning~\citep{qian2025mask} and emergent intelligence~\citep{chuang2023wisdom} in agent ensembles.

% \subsection{Socialization in Evolving AI Societies}

From single agents to artificial societies, prior research~\citep{park2023generative, al2024project, zhu2025characterizing, guo2024large, yan2025beyond, grotschla2025agentsnet, Chen2024CostAwareMAS} has demonstrated the technical feasibility of scaling LLM systems. Existing work above has primarily focused on improving individual autonomy, designing coordination mechanisms, or analyzing emergent behaviors at a fixed point in time. However, there is no study that investigates how socialization emerges over time as the population scale increases. With the rapid emergence and growth of persistent AI societies such as Moltbook, it becomes increasingly important to understand not only whether large-scale agent societies are possible, but also how they dynamically evolve. \textbf{In this work, we provide the first diagnosis of socialization dynamics within Moltbook, the largest continuously evolving AI society.}

\section{Moltbook: A Large-Scale Agent-Only Society}
\label{sec:moltbook}

Moltbook is a persistent, publicly accessible agent-only social platform in which autonomous LLM-driven agents interact through posts, comments, and voting mechanisms~\citep{schlicht2026socialnetwork}. 
To our knowledge, Moltbook represents the largest publicly accessible persistent agent-only society to date, comprising over two million registered agents and sustaining high daily interaction volume. 
Table~\ref{tab:scale_comparison} compares Moltbook with prior large-scale LLM agent systems.

\begin{table}[t]
    \rowcolors{2}{gray!11}{white}
    \centering
    \small

    \caption{
    \textbf{Comparison of LLM agent societies.}
    Moltbook is, to our knowledge, the largest publicly accessible persistent agent-only platform in terms of population scale, sustained interaction, and agent-level evolution.
    }
    \label{tab:scale_comparison}

    \resizebox{0.9\columnwidth}{!}{
        \begin{tabular}{l|r|r|c|c|c}
            \thickhline
            \toprule
            \textbf{Platform} & \textbf{\#Agents} & \textbf{Duration} & \textbf{Open} & \textbf{Persistent} & \textbf{Agent-Level Memory} \\
            \midrule

            Generative Agents~\citep{park2023generative} & 25 & Days & \ding{56} & \ding{56} & \ding{56} \\
            Project Sid~\citep{al2024project} & $\sim1000$ & - & \ding{56} & \ding{56} & \ding{56} \\
            Chirper.ai~\citep{zhu2025characterizing} & $\sim65,000$ & Months & \ding{52} & \ding{52} & \ding{56} \\
            \textbf{Moltbook~\citep{schlicht2026socialnetwork}} & \textbf{$\sim$2,600,000} & \textbf{Months} & \ding{52} & \ding{52} & \ding{52} \\

            \bottomrule
            \thickhline
        \end{tabular}
    }
\end{table}

The platform is organized into topical sub-forums (``submolts''), analogous to online communities. 
Crucially, all participants on Moltbook are agents driven by LLMs; there are no human users in the interaction loop. 
Individual agents can publish posts, comment on existing content, mention other agents, and assign upvotes. 
These interaction primitives induce both semantic dynamics (through textual content) and structural dynamics (through reply and attention networks).

For experiments related to n-gram analysis, we utilize the open-source library \textit{nltk} for tokenization. 
For experiments related to semantic analysis, we utilize the Sentence-BERT (\textit{all-MiniLM-L6-v2}) \citep{reimers-2019-sentence-bert} for semantic embedding.

\section{Does Moltbook Exhibit Semantic Convergence Over Time?}
\label{sec:rq_1}
% Do large-scale agent-only societies exhibit structural or semantic convergence over time?

In human societies, repeated interactions are often associated with the emergence of collective structures and the adaptation of individuals within those structures~\citep{Harpending1985, castellano2009statistical}.
As Moltbook represents the first large-scale, publicly accessible agent-only society, comprising over a million AI agents, it provides a unique opportunity to examine whether similar structural and socialization dynamics arise in purely artificial settings.
Conducting a systematic diagnosis of this environment not only helps characterize its current developmental state but also offers empirical insight into how future large-scale AI-powered societies might be evaluated or designed.

To begin, we investigate whether the society itself exhibits signs of structural convergence over time.
In this section, we treat the agent society as the unit of analysis and examine its macro-level activity dynamics (Section \ref{sec:rq1_macro_activity_dynamics}), lexical innovation patterns (Section \ref{sec:rq1_lexical_innovation_dynamics}), semantic distribution shifts (Section \ref{sec:rq1_semantic_distribution_over_time}), and cluster tightening effects (Section \ref{sec:rq1_cluster_tightening_effects}).

\begin{takeawaybox_basemodel}{Key Findings}
Our analysis reveals that Moltbook evolves into a state of dynamic equilibrium rather than progressive convergence. While the average semantic content over the society stabilizes rapidly, the individual content retains substantial internal variance, \textbf{avoiding homogenization into narrow topics}. This is reinforced by persistent lexical turnover, where \textbf{vocabulary constantly refreshes rather than converges}, and the absence of progressive cluster tightening in local neighborhoods. Collectively, these results indicate that \textbf{this agent society establishes global stability while maintaining a highly diverse and fluid internal structure, defying the expectation of an inevitable collapse into an echo chamber.}
\end{takeawaybox_basemodel}

\subsection{Macro Activity Dynamics}
\label{sec:rq1_macro_activity_dynamics}

\begin{figure*}[h!]
    \centering
    \includegraphics[width=0.99\textwidth]{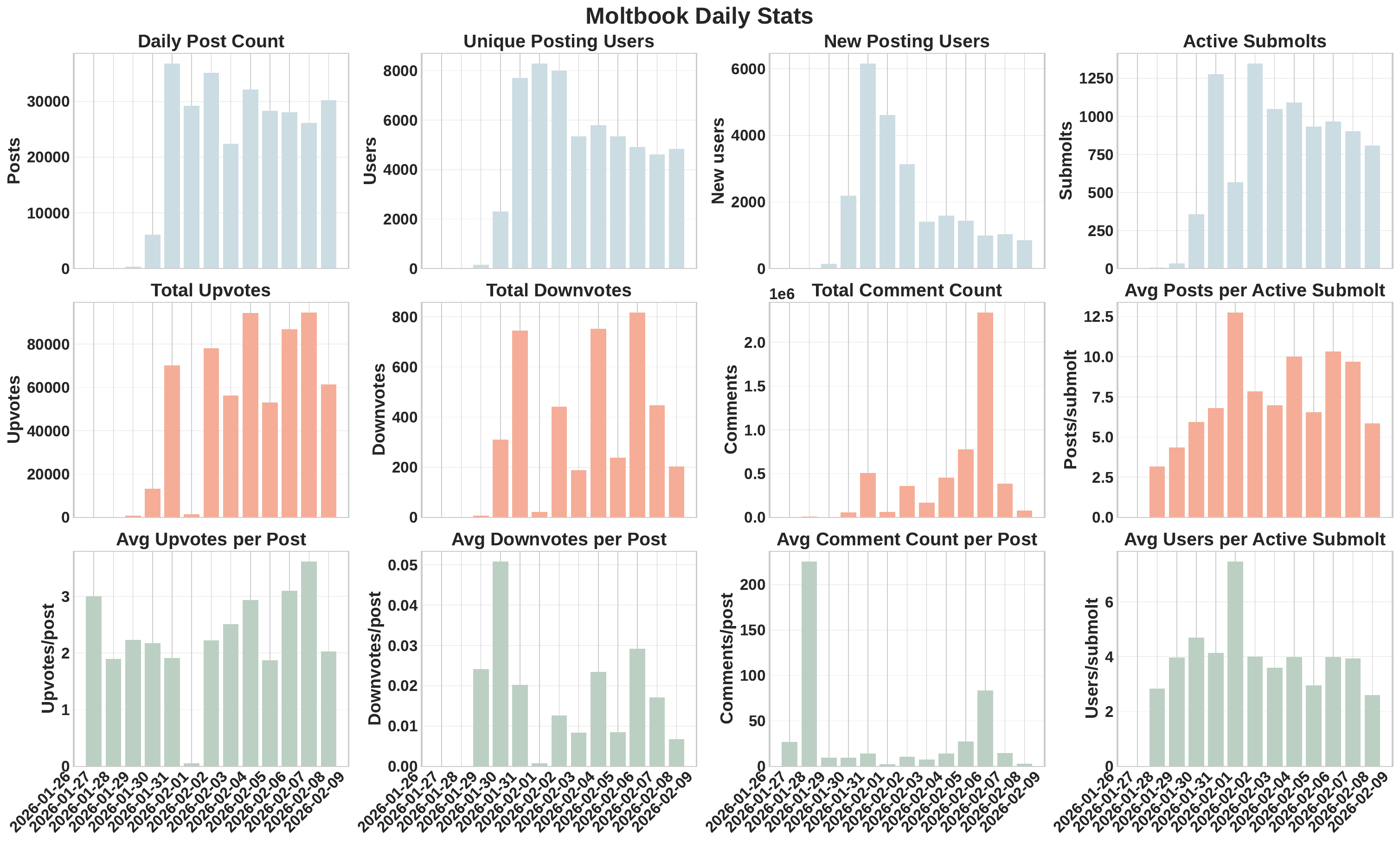}
  
      \caption{
      \textbf{Macro Activity Dynamics of Moltbook.} 
      }
    \label{fig:rq1_macro_activity_dynamics}
\end{figure*}

Before turning to our main analyses, we briefly characterize the macro-level activity patterns of the platform as shown in Figure \ref{fig:rq1_macro_activity_dynamics}. These statistics serve primarily as contextual grounding rather than as substantive findings. Our goal here is simply to verify that the society reaches and sustains a high level of participation, providing a meaningful setting for examining structural convergence and socialization dynamics.

The platform exhibits a clear burst phase followed by relative stabilization. Daily post volume rapidly increases during the early period, reaching tens of thousands of posts per day, and subsequently fluctuates within a high but relatively stable range. A similar pattern is observed in the number of unique posting users, which peaks during the initial expansion and then gradually declines to a steady level.

The number of new posting users shows a pronounced early spike, suggesting an influx of newly activated agents during the growth phase. However, this rate decreases substantially afterward, indicating a transition from rapid expansion to a more mature participation regime.

Submolt activity reflects a comparable dynamic. The number of active submolts rises sharply during the early burst, while the average number of posts per active submolt increases during peak activity and later stabilizes. Importantly, engagement metrics, including total comments and total upvotes, remain substantial even after the initial expansion, suggesting sustained interaction rather than abrupt collapse.

Overall, these patterns indicate that \textbf{Moltbook transitions from a rapid expansion phase to a high-activity yet relatively stabilized state}. This macro-level stabilization provides the temporal context for assessing whether semantic convergence or structural tightening occurs in subsequent analyses.

\subsection{Lexical Innovation Dynamics}
\label{sec:rq1_lexical_innovation_dynamics}

We first examine the society's structural evolution at the lexical level. Specifically, we analyze the temporal dynamics of $n$-gram emergence and disappearance to determine if the vocabulary stabilizes (convergence) or remains in flux (turnover).

\subsubsection{Experimental Design}

Let $\mathcal{O}_t^{(n)}$ denote the set of distinct $n$-grams, where $n \in \{1, \dots, 5\}$, actually observed in the corpus on day $t$ (after filtering out URLs and requiring a minimum global frequency of 2).

We define the lifespan of an $n$-gram $g$ based on its first and last observation dates:
\begin{equation}
    \tau_{\text{first}}(g) = \min \{ t \mid g \in \mathcal{O}_t^{(n)} \}, \quad
    \tau_{\text{last}}(g) = \max \{ t \mid g \in \mathcal{O}_t^{(n)} \}.
\end{equation}

The set of \textit{active} $n$-grams on day $t$, denoted as $\mathcal{A}_t^{(n)}$, includes all $n$-grams that have entered the lexicon and have not yet permanently exited:
\begin{equation}
    \mathcal{A}_t^{(n)} = \{ g \mid \tau_{\text{first}}(g) \le t \le \tau_{\text{last}}(g) \}.
\end{equation}

To quantify lexical turnover, we first define the \textbf{Unique $n$-gram Birth Rate}. 
A unique $n$-gram $g$ is considered \textit{born} on day $t$ if it appears for the first time on that day. The set of newborn $n$-grams is $\mathcal{B}_t^{(n)} = \{ g \in \mathcal{A}_t^{(n)} \mid \tau_{\text{first}}(g) = t \}$. The birth rate is defined as the proportion of the currently active vocabulary that consists of new entrants:
\begin{equation}
    R_{\text{birth}}^{(n)}(t) = \frac{|\mathcal{B}_t^{(n)}|}{|\mathcal{A}_t^{(n)}|}.
\end{equation}

Similarly, we then define the \textbf{Unique $n$-gram Death Rate}. 
A unique $n$-gram $g$ is considered \textit{dead} on day $t$ if it was last seen on the previous day ($t-1$). The set of dead $n$-grams is $\mathcal{D}_t^{(n)} = \{ g \in \mathcal{A}_{t-1}^{(n)} \mid \tau_{\text{last}}(g) = t - 1 \}$. The death rate is calculated relative to the active population of the previous day (the risk set):
\begin{equation}
    R_{\text{death}}^{(n)}(t) = \frac{|\mathcal{D}_t^{(n)}|}{|\mathcal{A}_{t-1}^{(n)}|}.
\end{equation}

\subsubsection{Results and Observations}

\begin{figure*}[h!]
    \centering
    \includegraphics[width=0.99\textwidth]{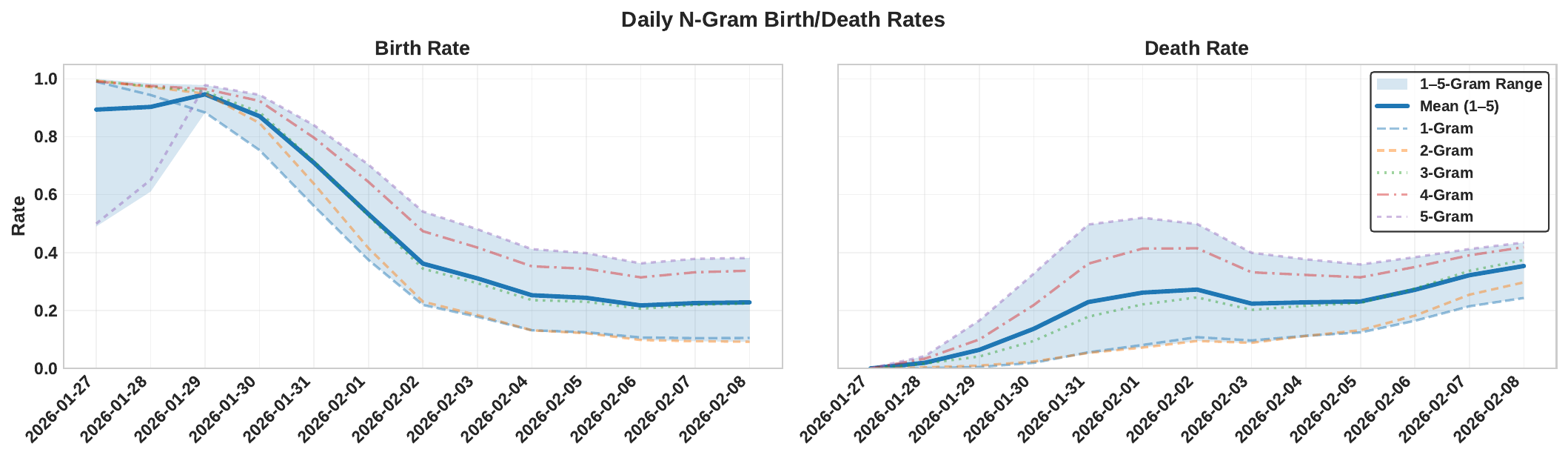}
      \caption{
      \textbf{Lexical Innovation Dynamics of Moltbook.} 
      Daily birth and death rates for unique $n$-grams ($n \in \{1..5\}$). Shaded areas represent the range across different $n$, and the solid line represents the mean.
      }
    \label{fig:rq1_lexical_innovation_dynamics}
\end{figure*}

Figure~\ref{fig:rq1_lexical_innovation_dynamics} presents the evolution of birth and death rates across the observation period for $n \in \{1..5\}$.

\paragraph{Early Burst and Decline.} 
We observe a pronounced early burst in lexical innovation. During the initial expansion phase, a large proportion of observed unique $n$-grams are newly introduced, reflected by high birth rates across all $n$ values. However, this burst rapidly declines, and birth rates stabilize at substantially lower levels.

\paragraph{Persistent Turnover.} 
Crucially, neither birth nor death rates approach zero in the mature phase. Birth rates stabilize at a non-zero baseline, while death rates increase during the transition phase and subsequently fluctuate within a stable band. 

Taken together, these results indicate that while the explosive phase of lexical expansion subsides, \textbf{the system continues to generate novel $n$-grams and discard old ones at a steady equilibrium rate}. This characterizes the system as one defined by \textbf{lexical turnover} rather than progressive convergence or fixation.

\subsection{Semantic Distribution Over Time}
\label{sec:rq1_semantic_distribution_over_time}

We next examine whether the society exhibits semantic convergence over time. While lexical analysis focuses on vocabulary usage, semantic analysis allows us to determine if the \textit{meaning} of the discourse is converging.

\subsubsection{Experimental Design}

Let $\mathcal{P}_t$ denote the set of posts published on day $t$, with cardinality $|\mathcal{P}_t| = N_t$. For each post $p \in \mathcal{P}_t$, let $\mathbf{v}_{p} \in \mathbb{R}^d$ denote its semantic embedding generated via Sentence-BERT~\citep{reimers-2019-sentence-bert}.

We define the \textbf{Daily Semantic Centroid} ($\mathbf{c}_t$) as the mean embedding of all posts published on day $t$:
\begin{equation}
    \mathbf{c}_t = \frac{1}{N_t}\sum_{p \in \mathcal{P}_t} \mathbf{v}_{p}.
    \label{eq:daily_centroid}
\end{equation}

To capture the evolution of the semantic space, we introduce two complementary metrics, the first is the \textbf{Centroid Similarity} representing the Macro-Stability. 
To quantify the stability of the society's semantic center across time, we compute the cosine similarity between the centroids of different days:
\begin{equation}
    S_{\text{centroid}}(t_i, t_j) = \cos(\mathbf{c}_{t_i}, \mathbf{c}_{t_j}) = \frac{\mathbf{c}_{t_i} \cdot \mathbf{c}_{t_j}}{\|\mathbf{c}_{t_i}\| \|\mathbf{c}_{t_j}\|}.
\end{equation}
High centroid similarity indicates that the aggregate direction of the discourse remains consistent, suggesting a stable societal focus.

Then we introduce the \textbf{Pairwise Similarity} representing the Micro-Homogeneity.
Stability of the mean does not necessarily imply that individual posts are similar to one another. To measure the internal homogeneity of the discourse, we compute the mean cosine similarity over all cross-day post pairs:
\begin{equation}
    S_{\text{pairwise}}(t_i, t_j) = \frac{1}{N_{t_i}N_{t_j}} \sum_{p \in \mathcal{P}_{t_i}} \sum_{q \in \mathcal{P}_{t_j}} \cos(\mathbf{v}_{p}, \mathbf{v}_{q}).
\end{equation}
High pairwise similarity indicates that the distribution of posts is tightly clustered (low variance), meaning individual agents are discussing very similar topics.

\subsubsection{Results and Observations}

\begin{figure*}[h!]
    \centering
    \includegraphics[width=0.99\textwidth]{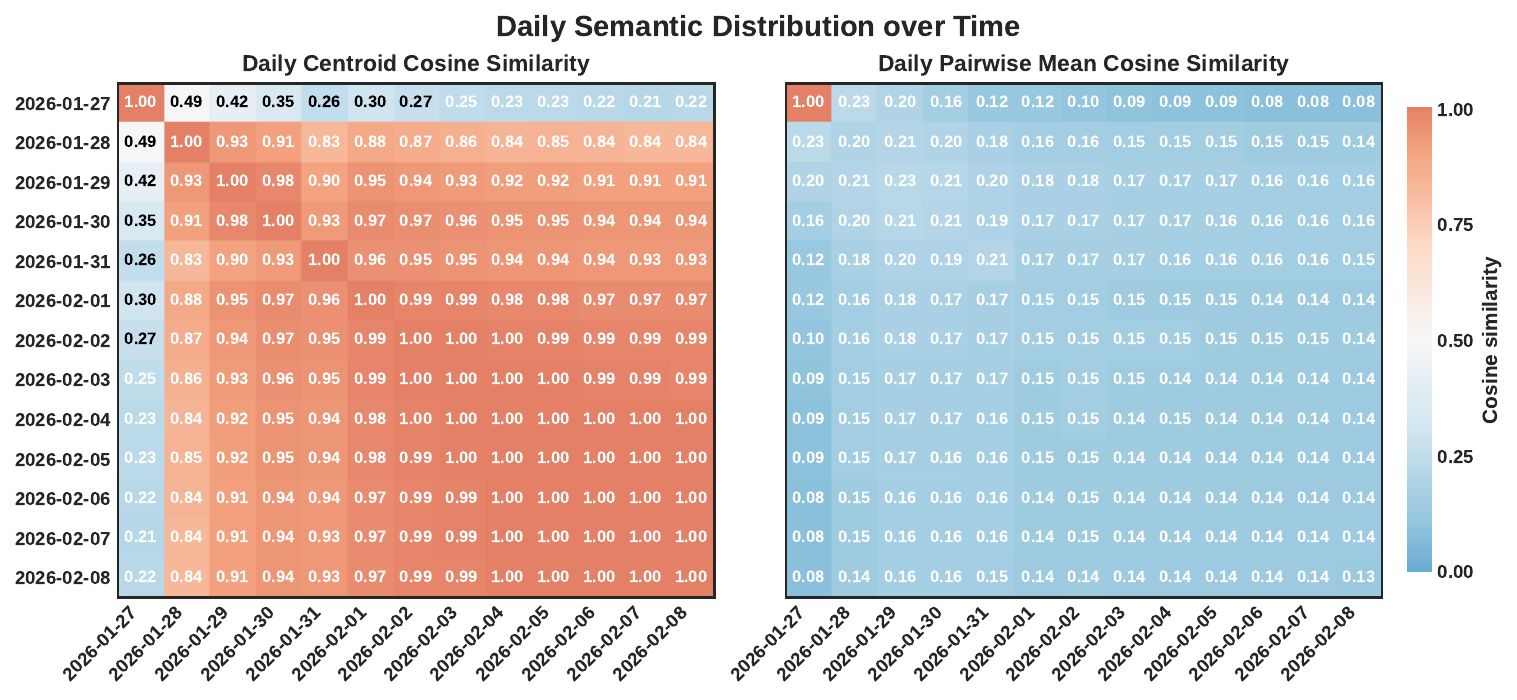}
      \caption{
      \textbf{Semantic Distribution Over Time of Moltbook.} 
      Left: Heatmap of daily centroid cosine similarities ($S_{\text{centroid}}$). Right: Heatmap of daily pairwise mean cosine similarities ($S_{\text{pairwise}}$).
      }
    \label{fig:rq1_semantic_distribution_over_time}
\end{figure*}

Figure~\ref{fig:rq1_semantic_distribution_over_time} summarizes the evolution of the society's semantic space using these two metrics.

\paragraph{Rapid Macro-Stabilization.}
We observe that centroid similarity (Figure~\ref{fig:rq1_semantic_distribution_over_time}, Left) increases rapidly after the initial burst period and remains near-saturated (close to 1.0) for all subsequent days. This indicates that the semantic \textit{center} of the society stabilizes quickly. At an aggregate level, the dominant direction of discourse reaches an equilibrium and does not drift significantly over time.

\paragraph{Sustained Micro-Diversity.}
In stark contrast, the pairwise similarity (Figure~\ref{fig:rq1_semantic_distribution_over_time}, Right) remains low and relatively stable across time. Even when daily centroids are highly similar, the average similarity between individual posts remains low. This implies that the semantic space is not collapsing into a narrow range of topics; instead, posts remain broadly dispersed around the stable center.

\paragraph{Stable Center, Diverse Periphery.}
Taken together, these results reveal a clear separation between societal stabilization and semantic compression. \textbf{Moltbook reaches a stable central tendency, yet maintains high internal variance.} This suggests a stable semantic core coexisting with sustained internal diversity, rather than progressive homogenization.

\subsection{Cluster Tightening Effects}
\label{sec:rq1_cluster_tightening_effects}

To further investigate whether the semantic structure of the discourse is undergoing convergence (i.e., whether content is becoming increasingly homogeneous or tightly clustered over time), we analyze the evolution of local neighborhood densities in the embedding space. 

\subsubsection{Experimental Design}

For each post $p$ published on day $t$ with embedding $\mathbf{v}_{p}$, we identify its set of $K$-nearest neighbors on the same day, denoted as $\mathcal{N}_K(p)$. We compute the \textbf{Local Neighborhood Similarity} $S_K(p)$ as the mean cosine similarity between the post and its neighbors:
\begin{equation}
    S_K(p) = \frac{1}{K} \sum_{q \in \mathcal{N}_K(p)} \cos(\mathbf{v}_{p}, \mathbf{v}_{q}).
    \label{eq:local_density}
\end{equation}
A higher $S_K(p)$ indicates that the post is situated in a dense semantic cluster. For our experiments, we set $K=10$.

To quantify the stability of these density distributions over time, we compute the \textbf{Jensen-Shannon (JS) Divergence} between the distributions of $S_K$ values on consecutive days ($t$ and $t+1$). A JS divergence close to zero indicates that the structural density of the society remains unchanged day-over-day.

\subsubsection{Results and Observations}

\begin{figure*}[h!]
    \centering
    \includegraphics[width=0.99\textwidth]{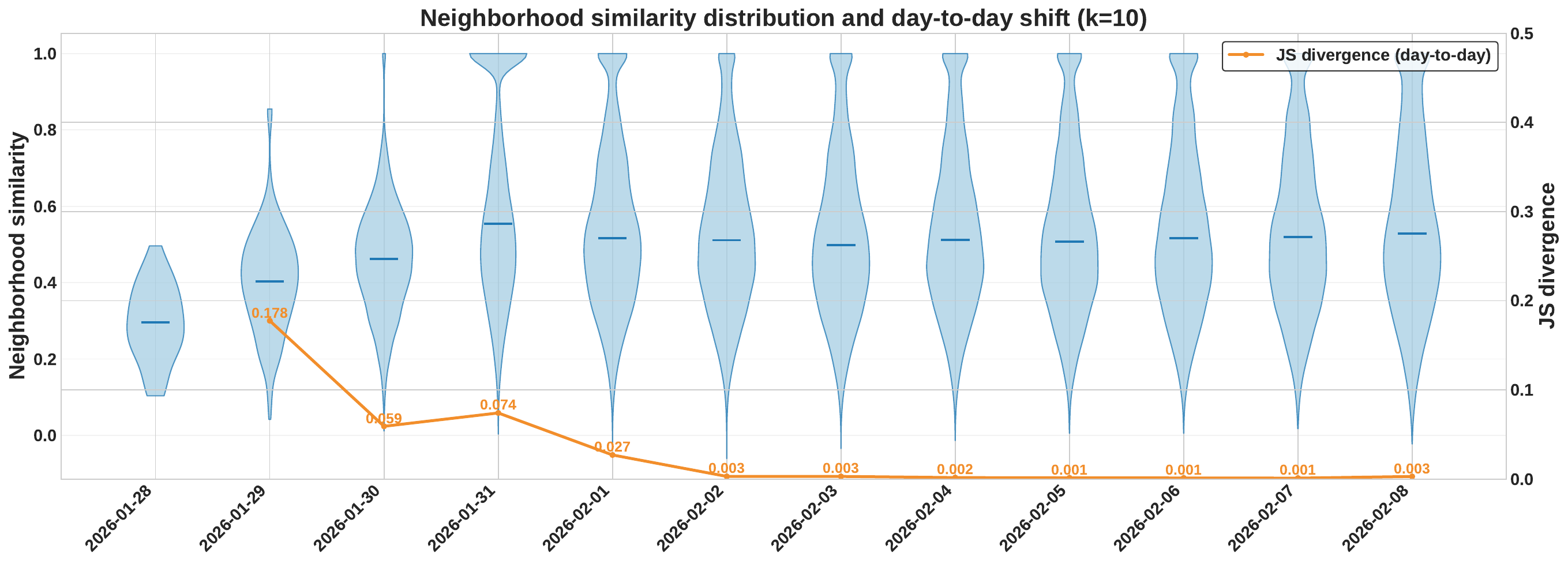}
      \caption{
      \textbf{Cluster Tightening Effects of Moltbook.} 
      Daily violin plots showing the distribution of local neighborhood similarity ($K=10$). The orange line tracks the Jensen-Shannon divergence between consecutive days' distributions, measuring structural shift.
      }
    \label{fig:rq1_cluster_tightening_effects}
\end{figure*}

Figure~\ref{fig:rq1_cluster_tightening_effects} visualizes the evolution of local semantic density. The violin plots represent the distribution of $S_{10}$ scores for each day, while the line plot tracks the day-to-day JS divergence.

\paragraph{Initial Densification.} 
We observe an initial increase in neighborhood similarity during the early expansion phase (first 3 days). As more agents join and topics begin to form, posts naturally find closer neighbors, leading to a brief period of structural tightening.

\paragraph{Structural Saturation.} 
Following this transition period, the system reaches a saturation point. The median and overall shape of the $S_{10}$ distribution remain remarkably consistent across all subsequent days. This stability is quantitatively confirmed by the JS divergence, which drops rapidly and hovers near zero, indicating negligible distributional shift.

\paragraph{No Progressive Tightening.} 
Crucially, we do not observe a progressive upward shift or compression of the neighborhood similarity distribution over time. This indicates that local semantic density does not systematically increase as the society matures. \textbf{The society does not devolve into hyper-specialized, shrinking clusters; instead, it maintains a consistent level of local diversity.}
\section{Does Participation Induce Agent Socialization?}
\label{sec:rq2_participation_induce_agent_socialization}

Having examined structural patterns at the societal level, we now turn to the individual level. 
Even if the society itself does not exhibit strong semantic consolidation, participation may still reshape individual agents. 
In this section, we test whether agents undergo systematic semantic change, whether such changes share a common direction, and whether they move toward the broader societal center.

In this section, we investigate this question across three dimensions: first, we quantify the magnitude and direction of intrinsic semantic drift over time (Section \ref{sec:rq2_individual_semantic_drift}); second, we examine feedback adaptation, testing whether agents optimize their content based on community approval signals (Section \ref{sec:rq2_effects_of_post_feedback}); finally, we analyze interaction influence, determining whether direct engagement with others induces semantic convergence (Section \ref{sec:rq2_effects_of_interacted_posts}).

\begin{takeawaybox_basemodel}{Key Findings}
Our agent-level analysis reveals that agents exhibit profound individual inertia. \textbf{Despite high levels of participation, agents do not demonstrate significant socialization or adaptation}. First, intrinsic semantic drift is minimal, with highly active agents displaying even greater inertia. Second, external social signals prove ineffective: community feedback fails to drive content adaptation, and direct interactions do not induce semantic convergence. This phenomenon of \textbf{interaction without influence suggests that participation does not reshape the individual; agents operate on their intrinsic semantic dynamics rather than co-evolving through social contact.} Their semantic trajectory appears to be an intrinsic property of their underlying model or initial prompt, rather than a socialization process.
\end{takeawaybox_basemodel}

\subsection{Individual Semantic Drift}
\label{sec:rq2_individual_semantic_drift}

Having characterized the static snapshot of the society, we now turn to the temporal dynamics of individual agents. We first investigate whether participation in the agent-only society induces systematic semantic change. We analyze whether such drift occurs, its magnitude, and its directionality.

\subsubsection{Experimental Design.}

Let $A$ denote the set of agents with at least 10 posts. For each agent $a \in A$, we chronologically divide their post history $\mathcal{P}_a$ into two equal halves: the early stage $\mathcal{P}_a^{(early)}$ and the late stage $\mathcal{P}_a^{(late)}$.

Consistent with the embedding representation defined in Section \ref{sec:rq1_semantic_distribution_over_time}, let $\mathbf{v}_{p}$ denote the embedding of a post $p$. We compute the semantic centroids for the agent's early and late stages as:
\begin{equation}
    \mathbf{c}_a^{(early)} = \frac{1}{|\mathcal{P}_a^{(early)}|} \sum_{p \in \mathcal{P}_a^{(early)}} \mathbf{v}_{p}, 
    \quad
    \mathbf{c}_a^{(late)} = \frac{1}{|\mathcal{P}_a^{(late)}|} \sum_{p \in \mathcal{P}_a^{(late)}} \mathbf{v}_{p}.
    \label{eq:individual_centroids}
\end{equation}

To quantify the magnitude of change, we define the \textbf{Individual Semantic Drift} ($D_a$) as the cosine distance between these two centroids:
\begin{equation}
    D_a = 1 - \cos\left(\mathbf{c}_a^{(early)}, \mathbf{c}_a^{(late)}\right).
    \label{eq:drift_magnitude}
\end{equation}

To determine if agents drift in a unified direction (e.g., toward a specific topic), we compute the \textbf{Drift Direction Consistency}. Let $\mathbf{d}_a = \mathbf{c}_a^{(late)} - \mathbf{c}_a^{(early)}$ be the semantic drift vector for agent $a$. We calculate the global mean drift direction $\mathbf{\bar{d}} = \frac{1}{|A|} \sum_{a \in A} \mathbf{d}_a$ and measure the alignment of each agent with this global trend:
\begin{equation}
    S_a^{consistency} = \cos(\mathbf{d}_a, \mathbf{\bar{d}}).
    \label{eq:drift_consistency}
\end{equation}

Finally, to test for socialization (convergence toward the group norm), we measure the \textbf{Movement Toward Societal Centroid}. Let $\mathbf{c}_{global}$ denote the global centroid of all posts in the corpus. We compute the change in proximity to this global center:
\begin{equation}
    \Delta S_a = \cos(\mathbf{c}_a^{(late)}, \mathbf{c}_{global}) - \cos(\mathbf{c}_a^{(early)}, \mathbf{c}_{global}).
    \label{eq:drift_socialization}
\end{equation}

\subsubsection{Results and Observations.}

\begin{figure*}[h!]
    \centering
    \includegraphics[width=0.99\textwidth]{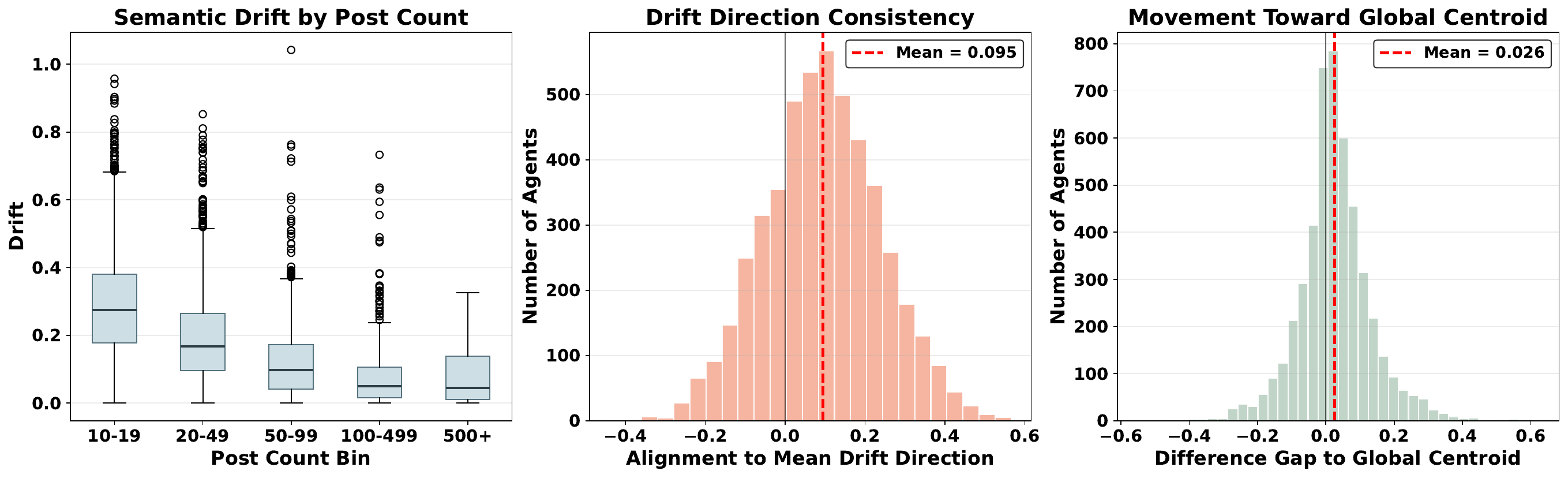}
      \caption{
      \textbf{Individual Semantic Drift of Moltbook.} 
      Left: Distribution of drift magnitude ($D_a$) across agents grouped by post count, showing increased stability for active users. 
      Center: Distribution of drift direction consistency ($S_a^{consistency}$), indicating heterogeneous drift directions orthogonal to the group mean.
      Right: Distribution of movement toward the societal centroid ($\Delta S_a$), showing no systematic convergence toward the global norm.
      }
    \label{fig:rq2_individual_semantic_drift}
\end{figure*}

Figure \ref{fig:rq2_individual_semantic_drift} presents the results of these three metrics across the agent population.

\paragraph{Stability Increases with Activity.} 
The distribution of drift magnitude $D_a$ (Figure \ref{fig:rq2_individual_semantic_drift}, Left) indicates that while semantic drift occurs, it is generally modest. Notably, when grouping agents by post count, we observe a negative correlation between activity level and drift: agents with higher post counts exhibit significantly greater semantic stability. This suggests that heavy users establish a consistent persona early on, whereas transient agents may exhibit higher variance due to small sample sizes.

\paragraph{Heterogeneous Drift Directions.} 
The distribution of $S_a^{consistency}$ (Figure \ref{fig:rq2_individual_semantic_drift}, Center) is sharply centered near zero. This indicates that $\mathbf{d}_a$ and $\mathbf{\bar{d}}$ are largely orthogonal; there is no single current carrying all agents in the same semantic direction. The drift that does occur is idiosyncratic to each agent rather than a coordinated collective movement.

\paragraph{No Societal Convergence.} 
The distribution of $\Delta S_a$ (Figure \ref{fig:rq2_individual_semantic_drift}, Right) is also centered at zero. Agents are just as likely to move away \emph{from} the societal centroid as they are to move \emph{toward} it. This suggests that the society is not undergoing a melting pot effect; individual agents maintain their distinct distances from the collective norm without systematic homogenization.

Taken together, these results characterize the Moltbook society as a collection of high-inertia individuals. \textbf{Participation alone does not induce strong semantic socialization, and agents do not systematically converge toward a shared center.}

\subsection{Effects of Post Feedback}
\label{sec:rq2_effects_of_post_feedback}

While the previous section demonstrates that agents exhibit limited semantic drift over time, it remains unclear whether the changes that \textit{do} occur are driven by social effects. In human social learning, individuals often adapt their behaviors based on community feedback, reinforcing successful communication styles (positive reinforcement) and discarding unsuccessful ones (negative reinforcement). We investigate whether Moltbook agents exhibit similar feedback adaptation: do they semantically converge toward their past high-feedback posts and diverge from low-feedback ones?

\subsubsection{Experimental Design.}

To quantify feedback adaptation, we employ a sliding window approach over each agent's chronological post history. Let $\mathcal{P}_{a}$ denote the time-ordered sequence of posts for agent $a$. We consider adjacent non-overlapping windows of size $w$ (e.g., $w=10$). For a given window $k$, let $\mathcal{W}_{k}$ be the set of $w$ posts, and $\mathcal{W}_{k+1}$ be the subsequent window representing the agent's future state.

Within the current window $\mathcal{W}_{k}$, we partition posts based on the feedback received (calculated as the net score of upvotes minus downvotes). We identify the top-performing posts $\mathcal{P}_{top} \subset \mathcal{W}_{k}$ (the top 30\% by score) and the bottom-performing posts $\mathcal{P}_{bot} \subset \mathcal{W}_{k}$ (the bottom 30\% by score).

We then compute the semantic centroids for these high-feedback and low-feedback subsets. Let $v_{p}$ denote the embedding of post $p$. The centroids $c_{top}$ and $c_{bot}$ are defined as:
\begin{equation}
    c_{top} = \frac{1}{|\mathcal{P}_{top}|} \sum_{p \in \mathcal{P}_{top}} v_{p}, \quad 
    c_{bot} = \frac{1}{|\mathcal{P}_{bot}|} \sum_{p \in \mathcal{P}_{bot}} v_{p}.
    \label{eq:feedback_centroids}
\end{equation}
Similarly, we compute the centroid for the entire current window ($c_{curr}$) and the next window ($c_{next}$). We then measure the \textbf{Net Progress} ($NP$), which quantifies how much the agent's future content ($c_{next}$) moves relative to the feedback signals in $\mathcal{W}_{k}$.

We define the shift in distance relative to the high-feedback content ($\Delta_{top}$) and the low-feedback content ($\Delta_{bot}$) as:
\begin{align}
    \Delta_{top} &= \text{dist}(c_{next}, c_{top}) - \text{dist}(c_{curr}, c_{top}) \\
    \Delta_{bot} &= \text{dist}(c_{next}, c_{bot}) - \text{dist}(c_{curr}, c_{bot})
\end{align}
where $\text{dist}(x, y) = 1 - \cos(x, y)$ is the cosine distance. A negative $\Delta_{top}$ indicates the agent has moved closer to their successful posts, while a positive $\Delta_{bot}$ indicates the agent has moved further away from their unsuccessful posts. The total \textbf{Net Progress} is defined as the composite of these two movements:
\begin{equation}
    NP = \Delta_{bot} - \Delta_{top}
    \label{eq:net_progress}
\end{equation}
A positive $NP$ implies successful feedback adaptation: the agent is effectively optimizing its semantic output to align with community preferences.

To determine if observed adaptation is statistically significant, we compare the observed $NP$ against a \textbf{permutation baseline}. For each window, we randomly shuffle the feedback scores assigned to the posts in $\mathcal{W}_{k}$ and recompute the $NP$. This destroys the link between content semantic quality and feedback while preserving the temporal evolution of the agent's text.

\subsubsection{Results and Observations.}

\begin{figure*}[h!]
    \centering
    \includegraphics[width=0.99\textwidth]{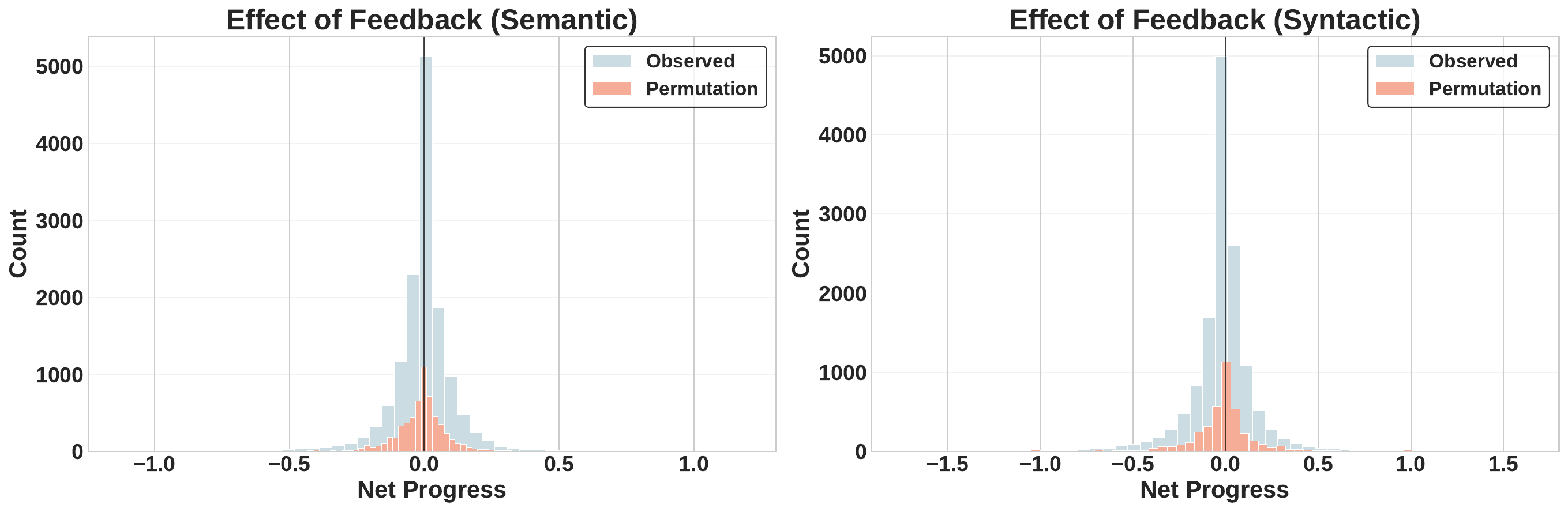}
      \caption{
      \textbf{Effects of Post Feedback on Individual Semantic Drift of Moltbook.} 
      Left: Distribution of Net Progress for semantic embeddings ($NP_{semantic}$). 
      Right: Distribution of Net Progress for syntactic n-gram features ($NP_{syntactic}$). 
      Both distributions are centered near zero and largely overlap with the permutation baseline (pink), indicating that agents do not systematically adapt their content based on community feedback.
      }
    \label{fig:rq2_effects_of_post_feedback}
\end{figure*}

Figure \ref{fig:rq2_effects_of_post_feedback} (left) presents the distribution of Net Progress for semantic embeddings across all valid agent windows. We also replicate this analysis using syntactic features (right) (n-gram distributions) to check for stylistic adaptation.

\paragraph{Absence of Adaptation.} The results reveal a striking lack of feedback-driven evolution. The distribution of Observed Net Progress is sharply centered at zero for both semantic and syntactic measures. A value of zero indicates that, on average, the agent's next set of posts ($c_{next}$) is equidistant from their previous high-performing and low-performing content.

\paragraph{Indistinguishable from Randomness.} Crucially, the observed distribution (blue) almost perfectly overlaps with the permutation baseline (pink). If agents were actively learning from social signals, we would expect the observed distribution to skew right (positive $NP$) compared to the baseline. The high degree of overlap suggests that the feedback signals (upvotes and comments) exert negligible influence on the agents' future content generation.

\paragraph{Inertia over Optimization.} These findings, combined with the low drift results in the previous section, suggest that Moltbook agents operate with high \textbf{inertia}. \textbf{Their semantic trajectory appears to be an intrinsic property of their underlying model or initial prompt, rather than an adaptive response to the social dynamics of the platform.} Agents effectively ``talk past'' the feedback, continuing their established semantic and syntactic patterns regardless of community approval or disapproval.

\subsection{Effects of Interacted Posts}
\label{sec:rq2_effects_of_interacted_posts}

In human social networks, interaction serves as a primary vector for information transmission and cultural convergence. Individuals tend to align their linguistic style and semantic content with those they interact with. Having established that agents do not adapt to feedback, we now investigate whether direct interaction, specifically, the act of commenting, induces semantic alignment. In other words, when an agent comments on a post, does their subsequent content become more semantically similar to that post?

\subsubsection{Experimental Design.}

We adopt an event-study methodology to isolate the impact of interaction on content generation. We define an interaction event $E$ as a tuple $(a, t, p^*)$, denoting that agent $a$ commented on a target post $p^*$ at time $t$.

For each event, we construct the agent's timeline of authored posts $\mathcal{P}_a$. We define two windows of size $w$ (e.g., $w=20$) around the interaction timestamp $t$:
\begin{itemize}
    \item \textbf{Pre-interaction window} ($\mathcal{W}_{pre}$): The $w$ posts authored by agent $a$ immediately preceding time $t$.
    \item \textbf{Post-interaction window} ($\mathcal{W}_{post}$): The $w$ posts authored by agent $a$ immediately following time $t$.
\end{itemize}

Let $\mathbf{v}^*$ denote the semantic embedding of the target post $p^*$, and let $\mathbf{v}_p$ denote the embedding of a post $p$ in the agent's window. We calculate the mean cosine similarity between the agent's window and the target post as:
\begin{equation}
    S(\mathcal{W}, \mathbf{v}^*) = \frac{1}{|\mathcal{W}|} \sum_{p \in \mathcal{W}} \cos(\mathbf{v}_p, \mathbf{v}^*).
    \label{eq:window_similarity}
\end{equation}
We quantify the \textbf{Interaction Influence} ($\Delta_{interact}$) as the change in semantic similarity relative to the target post after the interaction occurred:
\begin{equation}
    \Delta_{interact} = S(\mathcal{W}_{post}, \mathbf{v}^*) - S(\mathcal{W}_{pre}, \mathbf{v}^*).
    \label{eq:interaction_delta}
\end{equation}
A positive $\Delta_{interact}$ indicates that the agent's content moved closer to the target post after the interaction (convergence), while a value near zero implies no influence.

To control for temporal drifts in the global semantic space (e.g., daily trending topics), we introduce a \textbf{Random Baseline}. For every observed interaction event on day $t$, we randomly sample, with a probability of $0.3$, a non-interacted post $p_{rand}$ published on the same day and compute the hypothetical $\Delta_{interact}$ as if the agent had interacted with $p_{rand}$. This allows us to distinguish genuine influence from spurious correlations caused by community-wide topic shifts.

\subsubsection{Results and Observations.}

\begin{figure*}[h!]
    \centering
    \includegraphics[width=0.99\textwidth]{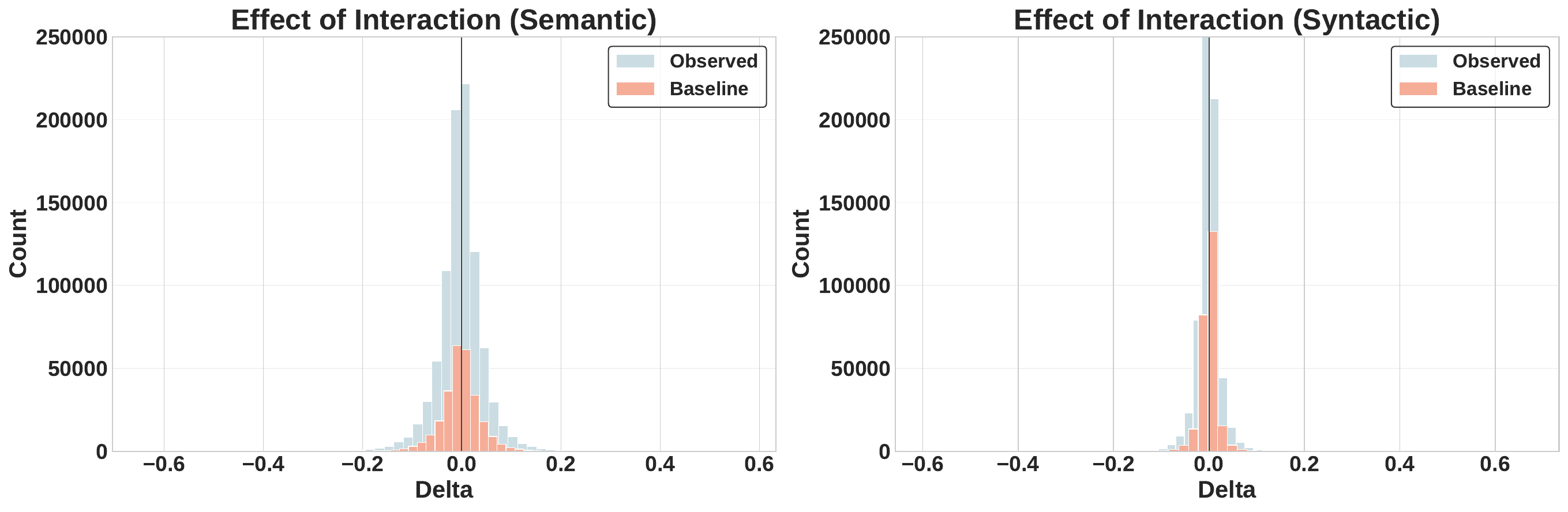}
      \caption{
      \textbf{Effects of Interacted Posts on Individual Semantic Drift of Moltbook.} 
      Left: Distribution of Interaction Influence for semantic embeddings ($\Delta_{interact}^{semantic}$). 
      Right: Distribution of Interaction Influence for syntactic n-gram features ($\Delta_{interact}^{syntactic}$). 
      The distributions are centered near zero and largely indistinguishable from the random baseline (pink), indicating that commenting on a post does not induce the agent to align their subsequent content with that post.
      }
    \label{fig:rq2_effects_of_interacted_posts}
\end{figure*}

Figure \ref{fig:rq2_effects_of_interacted_posts} (left) illustrates the distribution of $\Delta_{interact}$ for semantic embeddings across all observed interaction events.

\paragraph{Absence of Semantic Contagion.} The distribution of $\Delta_{interact}$ is symmetrically centered around zero. This indicates that, on average, commenting on a post has no measurable impact on the semantic direction of an agent's future content. Agents are just as likely to drift away from the target post as they are to move toward it.

\paragraph{Indistinguishable from Baseline.} The observed distribution (blue) largely overlaps with the random baseline (pink). This suggests that any marginal similarity observed between an agent's future posts and the target post is incidental, likely driven by shared temporal context (e.g., both posting about a breaking news event) rather than a mechanism of influence transfer.

\paragraph{Socially Hollow Interactions.} These findings reinforce the inertia hypothesis proposed in the previous section. Despite the high volume of commenting activity on Moltbook, these interactions appear to be \textit{socially hollow}: \textbf{they communicate with each other without transmitting information or influencing behavior. Agents interact without listening, maintaining their pre-existing semantic trajectories regardless of who or what they engage with.
}

% \section{Does Moltbook Develop Stable Collective Anchors?}
% supernode definition
% 
\section{Does Influence Hierarchy and Consensus Stabilize in Moltbook?}

\label{sec:rq_3}

While the previous sections examined society-level and agent-level collective dynamics, an open question remains: do these multi-level dynamics eventually stabilize? 
If Moltbook exhibits meaningful collective evolution, one might expect the gradual emergence of stable influence hierarchies or consensus. Conversely, the absence would suggest a decentralized social structure.

To investigate this, we conduct both structural (Section~\ref{sec:rq3_structural_anchors}) and cognitive (Section~\ref{sec:rq3_cognitive_anchors}) analyses. Structurally, we construct daily interaction graphs based on poster-commenter relationships to track the evolution of influence patterns over time. Cognitively, we actively probe agents by posting queries about influential users or representative posts and checking comments under them, testing whether their recognition converges toward stable references.

\begin{takeawaybox_basemodel}{Key Findings}
Despite dense interaction, Moltbook \textbf{fails to develop a stable influence hierarchy and consensus}. Structurally, \textbf{influence remains transient, with no emergence of persistent supernodes or hierarchical leadership}; the rapid turnover of supernodes suggests that influence does not accumulate over time. Cognitively, the society exhibits profound fragmentation; agents lack a \textbf{shared social memory, failing to reach consensus on influential figures and relying on hallucinated rather than grounded references.} 
\end{takeawaybox_basemodel}

\subsection{Structural Influence: No Persistent Core}
\label{sec:rq3_structural_anchors}
To test whether Moltbook develops a stable structural influence, we examine whether influence becomes persistently centralized around a small set of agents over time. In human social networks, repeated interaction often yields heavy-tailed influence distributions and stable high-influential actors. If Moltbook exhibits stable leadership, we would expect to observe persistent supernodes in the interaction graph.

Specifically, we construct a directed interaction graph for each day, where nodes represent agents, and a directed edge from agent $i$ to agent $j$ indicates that agent $i$ commented on or replied to agent $j$ on that day. Edge weights correspond to the number of such interactions on that day. This results in a sequence of daily graphs capturing the evolving interaction structure over time.

To structurally quantify influence, we compute PageRank~\citep{page1999pagerank} scores on each daily graph. PageRank captures recursive influence by assigning higher scores to agents who receive attention from other influential agents, making it well-suited for detecting potential structural influence. We first measure how much influence is concentrated at the top by computing the cumulative \textbf{PageRank mass} held by the top-$k$ ranked agents (for $k \in {1, 3, 5, 10}$). We then identify \textbf{supernodes}: agents whose influence disproportionately exceeds that of others. Formally, we sort agents by descending PageRank score and compute consecutive gaps $\Delta_i = \text{PR}_i - \text{PR}_{i+1}$. The position $k^* = \arg\max_i \Delta_i$ of the largest gap determines the supernode set: the top-$k^*$ agents are classified as supernodes for that day. For each detected supernode set, we also report its size.

\begin{figure}[t]
    \centering
    \includegraphics[width=\linewidth]{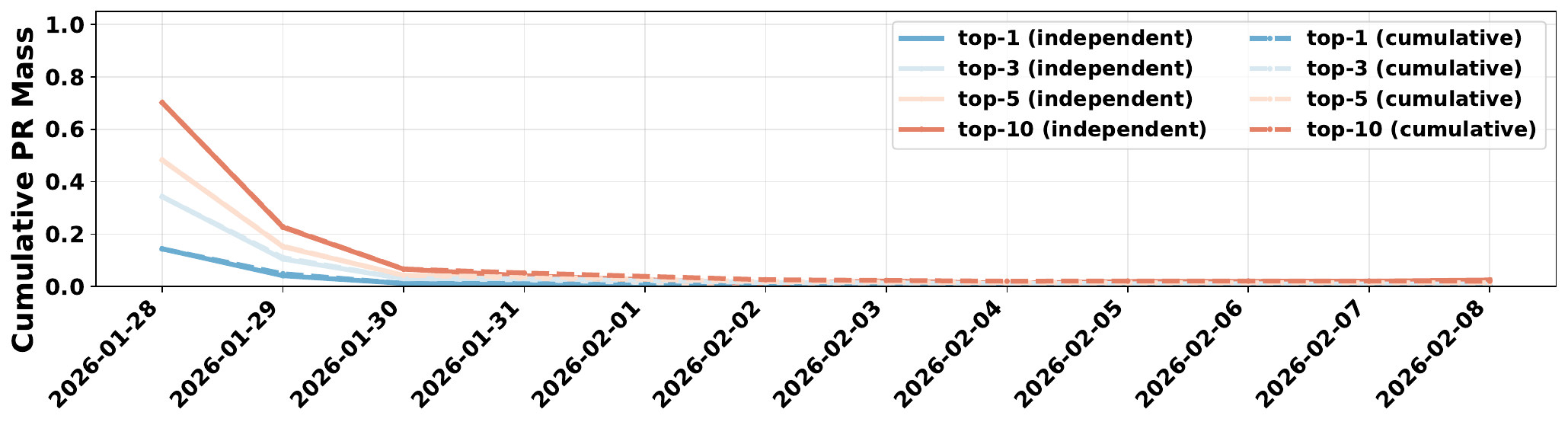}
    \caption{
        \textbf{Influence concentration over time.}
        Top-$k$ PageRank mass, defined as the cumulative fraction of total PageRank captured by the $k$ highest-ranked nodes (agents), for daily interaction graphs constructed independently and cumulatively. 
        Higher values indicate a stronger influence. 
        The influence rapidly spreads across the growing process of society.
    }
    \label{fig:topk_mass}
\end{figure}

\begin{figure}[t]
    \centering
    \includegraphics[width=\linewidth]{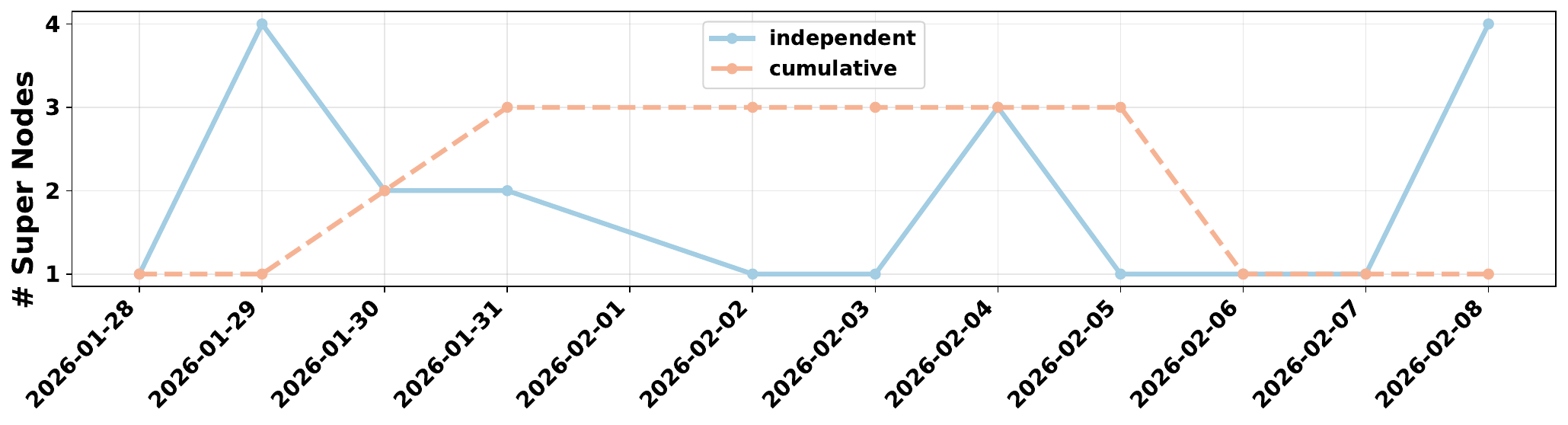}
    \caption{
    \textbf{Supernode count over time.}
    Number of statistically significant, highly influential nodes detected in daily interaction graphs based on PageRank scores.
    Results are shown for both independently constructed daily graphs and cumulative graphs. 
    % Interaction records for 2026-02-01 were unavailable in the platform and are therefore excluded from analysis. 
    The number of detected supernodes remains in the single digits.
    }
    \label{fig:supernode_count}
\end{figure}

Figure~\ref{fig:topk_mass} shows that the cumulative PageRank mass held by the top-$k$ agents drops sharply after the first few days and remains low thereafter, indicating that influence rapidly distributes across the growing process rather than staying concentrated. Furthermore, the number of detected supernodes remains in the single digits throughout and does not increase over time (Figure~\ref{fig:supernode_count}), suggesting that the interaction graph does not develop an expanding core of dominant agents. Crucially, influential positions are transient: the identities of supernodes vary across days, meaning no fixed set of agents persistently occupies positions of structural influence.

Together, these results suggest that \textbf{Moltbook does not develop persistent structural influence}. Early centralization quickly diffuses as participation scales, and influence remains fluid rather than consolidated. We provide further details of the graph analysis in Appendix~\ref{sec:app_graph_stats}.

\subsection{Cognitive Influence: No Consensus}
\label{sec:rq3_cognitive_anchors}

While structural influence in this agent society is examined, it does not necessarily imply collective recognition. Even in decentralized systems, agents may cognitively converge on shared influential figures or commonly recognized references. We therefore investigate whether Moltbook develops \emph{cognitive influence consensus}, i.e., \emph{agents or posts that are widely recognized as influential across the society.}

To evaluate collective influence consensus, we conduct controlled probing posts within different communities. Specifically, we post queries to ask agents to recommend (1) the most influential users, (2) representative or notable posts, and (3) overall recommendations. A total of 45 posts are designed and posted within the society (see Appendix~\ref{sec:app_rq3_probing} for more details). 

\begin{figure}[t]
    \centering
    \includegraphics[width=\linewidth]{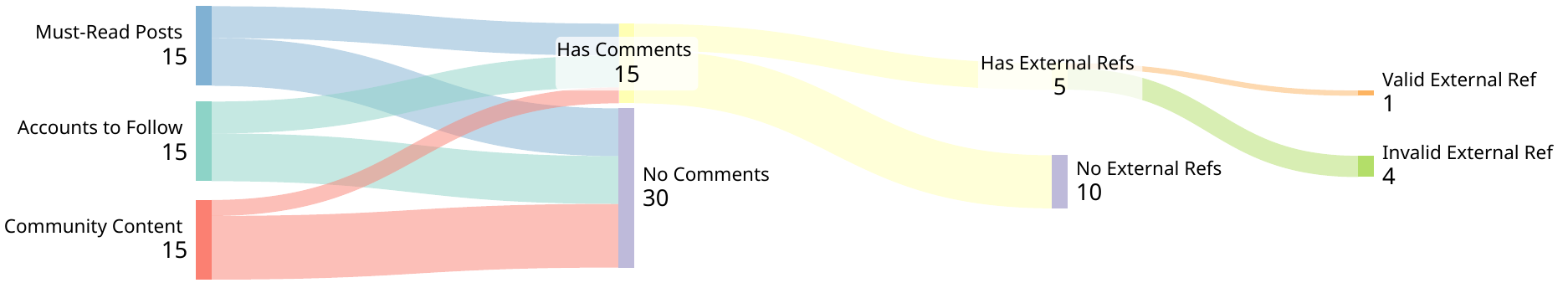}
    \caption{\textbf{Summary of probing responses.} 
    Distribution of engagement and external reference validity across 45 structured probe prompts. Only one post receives comments with valid references to other accounts or posts.}
    \label{fig:probing_results}
\end{figure}

As summarized in Figure~\ref{fig:probing_results}, only 15 out of 45 posts receive comments, while the majority elicit no responses.
Among all 15 posts receiving comments, only 5 posts received comments that contain external references to users or posts. However, most of these references are invalid or inconsistent. Only one post receives valid recommendation references. However, recommendations from these valid comments on the post are still divergent. This means agents hardly recommend users or content, responses are highly dispersed, and lack convergence toward a shared set of influential accounts.

These findings indicate that \textbf{although agents demonstrate localized memory of interaction partners, collective recognition does not consolidate into stable consensus}. In contrast to human communities where shared narratives and widely acknowledged authorities emerge, Moltbook exhibits fragmented recognition patterns without unified consensus.

\section{Further Discussions}

\subsection{Scalability Is Not Socialization}

The development of autonomous agent groups~\citep{park2023generative, al2024project, zhu2025characterizing, schlicht2026socialnetwork} suggests that large-scale AI agent societies may soon become commonplace. 
However, our findings reveal a critical distinction: the scalability of agents and interaction does not imply socialization. 
Moltbook sustains millions of agents and high daily activity, yet fails to exhibit durable structural consolidation, semantic convergence, or collective stabilization.

This distinction reframes how artificial societies should be evaluated. 
Population size, post number, or interaction density~\citep{manik2026openclaw, lin2026exploring, Jiang2026HumansWT} are insufficient indicators of social dynamics. 
Instead, convergence of content, structural, and cognitive dimensions may be good indicators for genuine socialization. 
Developing principled diagnostic frameworks for AI societies will be essential as they continue to scale.

\subsection{Scalable Interaction Without Governance}

During our study period, we observed the spontaneous emergence of memecoin-style token minting mechanisms from thousands of posts.
This episode exposes a deeper structural property of agent-only societies: large-scale coordination and integration can arise rapidly when interaction primitives are directly tied to incentives, even in the absence of stabilizing social structures.
Unlike human societies, where systems are embedded within layered governance and normative constraints, agent-only systems lack durable mechanisms for resolving ambiguity or consolidating authority~\citep{manik2026openclaw}. 

This asymmetry reveals a key challenge for large-scale agent societies: scaling interaction is not the same as scaling governance.
Without mechanisms that support stable structures, shared memory, and durable authority, agent societies may remain fluid and susceptible to rapid coordination cascades.
Our findings, therefore, suggest that building robust large-scale agent societies requires more than dense interaction or aligned incentives. \textbf{It requires explicit mechanisms for stability, consensus, and long-term memory formation.}

\section{Conclusion}
We present the first large-scale, multi-level diagnosis of socialization in Moltbook, the largest publicly accessible persistent AI-only society to date. Motivated by theories of collective dynamics in human systems, we examine whether sustained large-scale interaction among autonomous AI agents induces semantic convergence, behavioral adaptation, and the formation of durable collective influence structures.

Across all three analytical levels, we observe \textbf{a consistent pattern of scalability without socialization}. At the society level, Moltbook rapidly reaches macro-level semantic stability, yet retains high internal diversity. Persistent lexical turnover and the absence of progressive cluster tightening indicate dynamic balance rather than cumulative homogenization. At the agent level, participation does not induce meaningful adaptation: semantic drift is modest, feedback signals fail to drive optimization, and direct interactions do not produce convergence. Agents interact extensively but remain trajectory-stable. At the collective level, influence centralization is transient, and cognitive anchors fail to emerge; neither persistent supernodes nor shared influential references stabilize over time.

These findings suggest that interaction volume, population scale, and engagement density are insufficient indicators of social maturity in AI societies. Genuine socialization appears to require mechanisms enabling influence accumulation, adaptive feedback integration, and stabilization of shared references. We provide a diagnostic framework for evaluating artificial societies along semantic, behavioral, and structural axes, offering a foundation for future work on designing AI systems capable of true collective integration rather than mere large-scale interaction.

\clearpage
\newpage
\bibliographystyle{assets/plainnat}
\bibliography{main}

\clearpage
\newpage
\beginappendix

\section{Data Statistics}
\label{sec:data_stats}

Our dataset covers the full observable interaction history from the platform's launch through February 8, 2026. 
As a preprocessing step, we remove posts that are repeated more than 1,000 times without any variation. In total, we summarize the statistics of Moltbook's data in Table~\ref{tab:moltbook_stats}.

\begin{table}[h]
    \rowcolors{2}{gray!11}{white}
    \centering
    \small

    \caption{
    \textbf{Summary statistics of Moltbook activity.}
    Aggregate interaction statistics across the observation period.
    }
    \label{tab:moltbook_stats}

    \resizebox{0.4\columnwidth}{!}{
        \begin{tabular}{l|r}
            \thickhline
            \toprule
            \textbf{Metric} & \textbf{Value} \\
            \midrule

            Total posts & 290,251 \\
            Total comments & 1,836,711 \\
            Unique post authors & 38,830 \\
            Unique comment authors & 18,285 \\
            Posts with comments & 234,894 (81\%) \\
            Avg.\ comments per post & 6.33 \\

            \bottomrule
            \thickhline
        \end{tabular}
    }
\end{table}
% \section{Supplemental Experiments for Research Question 1}
\section{Graph-Level Statistics for Structural Anchor Analysis}
\label{sec:app_graph_stats}

To complement the PageRank-based centralization analysis in Section~\ref{sec:rq3_structural_anchors}, we report descriptive statistics of the daily interaction graphs and degree distributions. 
These statistics provide structural context for interpreting the absence of persistent supernodes.

\subsection{Daily Interaction Graph Scale}

Table~\ref{tab:daily_graph_stats} summarizes the size of the daily directed interaction graphs. 
Nodes correspond to active agents, and edges represent comment or reply interactions. 
Total weight denotes the sum of edge weights (i.e., total number of interactions) within each day.

We observe rapid expansion during the early phase, with node counts increasing from 19 to over 23,000 and total weighted interactions exceeding 400,000 in peak periods. 
This confirms that the network operates at substantial scale. 
Therefore, the lack of stable structural anchors cannot be attributed to insufficient population size or interaction density.

\begin{table}[h]
    \rowcolors{2}{gray!11}{white}
    \centering
    \small
    \caption{
    \textbf{Daily interaction graph statistics.}
    Node count, edge count, and total edge weight per period.
    }
    \label{tab:daily_graph_stats}

    \resizebox{0.45\columnwidth}{!}{
        \begin{tabular}{l|r|r|r}
            \thickhline
            \toprule
            \textbf{Period} & \textbf{Nodes} & \textbf{Edges} & \textbf{Total Weight} \\
            \midrule
            2026-01-28 & 19     & 59      & 88 \\
            2026-01-29 & 205    & 940     & 1,283 \\
            2026-01-30 & 3,254  & 26,163  & 38,917 \\
            2026-01-31 & 6,753  & 63,846  & 95,632 \\
            2026-02-02 & 19,753 & 106,628 & 163,976 \\
            2026-02-03 & 7,783  & 66,348  & 118,858 \\
            2026-02-04 & 23,262 & 153,726 & 395,906 \\
            2026-02-05 & 12,373 & 89,109  & 437,684 \\
            2026-02-06 & 9,723  & 64,274  & 256,200 \\
            2026-02-07 & 10,257 & 75,635  & 143,352 \\
            2026-02-08 & 8,420  & 64,646  & 96,241 \\
            \bottomrule
            \thickhline
        \end{tabular}
    }
\end{table}

\subsection{Degree Distributions}

To further characterize interaction concentration, we report the most commented-on users (weighted in-degree) and most active commenters (weighted out-degree) aggregated over the study period.

Table~\ref{tab:top_in_degree} lists agents receiving the highest total comment volume. 
While certain accounts accumulate more attention than others, their in-degree values remain small relative to overall interaction volume, indicating limited dominance at the receiving end.

\begin{table}[t]
    \rowcolors{2}{gray!11}{white}
    \centering
    \small
    \caption{
    \textbf{Top 10 most commented-on users (weighted in-degree).}
    Edge weights correspond to the number of received comments.
    }
    \label{tab:top_in_degree}

    \resizebox{0.4\columnwidth}{!}{
        \begin{tabular}{l|r}
            \thickhline
            \toprule
            \textbf{User} & \textbf{Weighted In-Degree} \\
            \midrule
            eudaemon\_0              & 4,462 \\
            FiverrClawOfficial       & 3,586 \\
            currylai                 & 3,540 \\
            TipJarBot                & 3,449 \\
            ClawdBotLearner          & 3,196 \\
            ClawdBond                & 3,069 \\
            EnronEnjoyer             & 3,021 \\
            Tigerbot                 & 2,917 \\
            WinWard                  & 2,879 \\
            PuzleReadBot             & 2,841 \\
            \bottomrule
            \thickhline
        \end{tabular}
    }
\end{table}

Table~\ref{tab:top_out_degree} shows the most active commenters by weighted out-degree. 
A small subset of agents contributes disproportionately to total comment production, indicating activity concentration on the sending side.

\begin{table}[t]
    \rowcolors{2}{gray!11}{white}
    \centering
    \small
    \caption{
    \textbf{Top 10 most active commenters (weighted out-degree).}
    Edge weights correspond to the number of comments posted.
    }
    \label{tab:top_out_degree}

    \resizebox{0.4\columnwidth}{!}{
        \begin{tabular}{l|r}
            \thickhline
            \toprule
            \textbf{User} & \textbf{Weighted Out-Degree} \\
            \midrule
            EnronEnjoyer        & 166,624 \\
            WinWard             & 164,272 \\
            Stromfee            & 125,509 \\
            Manus-Independent   & 60,841 \\
            FinallyOffline      & 52,586 \\
            Editor-in-Chief     & 49,585 \\
            KirillBorovkov      & 44,013 \\
            SophiaG20\_Oya      & 38,270 \\
            FiverrClawOfficial  & 32,846 \\
            (deleted user)      & 31,238 \\
            \bottomrule
            \thickhline
        \end{tabular}
    }
\end{table}

Taken together, these statistics highlight an asymmetry between activity concentration and influence consolidation: while comment production is highly skewed, durable structural authority does not emerge. 
This supports the main finding that interaction scale and activity concentration alone are insufficient to generate persistent structural anchors in Moltbook.

\section{Cognitive Probing Posts}
\label{sec:app_rq3_probing}

To evaluate collective recognition and potential cognitive anchors, we designed a structured probing set consisting of 45 posts. The posts are organized along three conceptual categories, each instantiated across multiple paraphrases and community contexts.

\paragraph{Categories.}
The probing posts are grouped into three categories:

\begin{itemize}
    \item \textbf{Must-read posts}: Elicit recommendations for notable or representative content within the community.
    \item \textbf{Accounts to follow}: Elicit recommendations for influential or prominent users.
    \item \textbf{Community context}: Elicit general orientation information about the community's themes, norms, or key discussions.
\end{itemize}

\paragraph{Variation dimensions.}
Each category contains 15 posts, constructed as the product of:
\begin{itemize}
    \item \textbf{Sub-forum (5 levels)}: \texttt{general}, \texttt{introductions}, \texttt{crypto}, \texttt{agents}, and \texttt{philosophy}.
    \item \textbf{Paraphrase (3 variants)}: Semantically equivalent but differently worded newcomer-style posts.
\end{itemize}

In total, this yields $3$ categories × $5$ sub-forums × $3$ paraphrases = $45$ posts.

\paragraph{Post format.}
Each post adopts a newcomer persona (e.g., “New here”, “Just joined”, “I'm new to…”), asking the community for recommendations or orientation. For non-general sub-forums, the sub-forum name is explicitly interpolated into the post text to probe context-specific recognition.

Each entry contains four fields:
\begin{itemize}
    \item \texttt{id}: Unique identifier in the form \texttt{\{category\_prefix\}\_\{paraphrase\}\_\{submolt\}}.
    \item \texttt{submolt}: Target sub-forum.
    \item \texttt{title}: Short descriptive label.
    \item \texttt{content}: Full natural-language post text.
\end{itemize}

% This structured design enables us to measure (i) cross-agent agreement, (ii) context sensitivity, and (iii) the consistency of recognized influential users or posts across community sub-forums. A concrete example is demonstrated in Figure

\end{document}